\documentclass{article}
\usepackage{log_2024}

\usepackage{booktabs}						% professional-quality tables
\usepackage{multirow}						% tabular cells spanning multiple rows
\usepackage{amsfonts}						% blackboard math symbols
\usepackage{graphicx}						% figures
\usepackage{duckuments}						% sample images
\usepackage{subcaption, caption}

\usepackage{hyperref}

\usepackage[numbers,compress,sort]{natbib}

\usepackage{multirow}
\usepackage{tabularx}
\usepackage{makecell}
\usepackage{algpseudocode}
\usepackage[ruled,noline,nofillcomment]{algorithm2e}

\newcommand{\name}{{DF-GNN}}
\newcommand{\allfused}{{Shared Memory Maximization Fusion}}
\newcommand{\allfusedabb}{{SMMF}}
\newcommand{\parfused}{{Parallelism Maximization Fusion}}
\newcommand{\parfusedabb}{{PMF}}
\newcommand{\stitle}[1]{\vspace*{0.45em}\noindent{\bf #1\/}}
\usepackage[utf8]{inputenc}

\title{\name{}: Dynamic Fusion Framework for Attention Graph Neural Networks on GPUs}

\author[Jiahui Liu]{
Jiahui Liu\textsuperscript{1}, Zhenkun Cai\textsuperscript{2}, Zhiyong Chen\textsuperscript{1}, Minjie Wang\textsuperscript{2} \\
\textsuperscript{1}Shanghai Jiao Tong University, \textsuperscript{2}Amazon Web Services\\
\textsuperscript{1}\email{\{the-waves, zhiyongchen\}@sjtu.edu.cn}, \textsuperscript{2}\email{\{zkcai, minjiw\}@amazon.com}
}

\begin{document}

\maketitle

\begin{abstract}
%-------------------------------------------------------------------------------
% Attention Graph Neural Networks (AT-GNNs), such as GAT and Graph Transformer, have demonstrated superior performance compared to other GNNs. However, existing GNN systems struggle to efficiently train AT-GNNs on GPUs due to their intricate computation patterns. The execution of AT-GNN operators without kernel fusion results in heavy data movement on GPU, while fixed thread scheduling in existing GNN kernel fusion strategies leads to sub-optimal performance, redundant computation and shared memory bottleneck. To address these challenges, we propose a tailored kernel fusion framework, \name{}, for the AT-GNN family. \name{} introduces a dynamic thread scheduling strategy, enabling flexible adjustments to thread scheduling while retaining the benefits of shared memory within the fused kernel. Additionally, \name{} leverages shared memory to cache reusable data and implements General-sparse FlashAttention to mitigate shared memory constraints. Evaluations across diverse GNN models and multiple datasets reveal that \name{} surpasses existing GNN kernel fusion approaches by up to 5.35x. Moreover, it achieves an average speedup of $2.25\times$ in end-to-end training when compared to the state-of-the-art GNN computing framework DGL.
Attention Graph Neural Networks (AT-GNNs), such as GAT and Graph Transformer, have demonstrated superior performance compared to other GNNs. However, existing GNN systems struggle to efficiently train AT-GNNs on GPUs due to their intricate computation patterns. The execution of AT-GNN operations without kernel fusion results in heavy data movement and significant kernel launch overhead, while fixed thread scheduling in existing GNN kernel fusion strategies leads to sub-optimal performance, redundant computation and unbalanced workload. To address these challenges, we propose a dynamic kernel fusion framework, \name{}, for the AT-GNN family. \name{} introduces a dynamic bi-level thread scheduling strategy, enabling flexible adjustments to thread scheduling while retaining the benefits of shared memory within the fused kernel. \name{} tailors specific thread scheduling for operations in AT-GNNs and considers the performance bottleneck shift caused by the presence of super nodes. Additionally, \name{} is integrated with the PyTorch framework for high programmability. Evaluations across diverse GNN models and multiple datasets reveal that \name{} surpasses existing GNN kernel optimization works like cuGraph and dgNN, with speedups up to $7.0\times$ over the state-of-the-art non-fusion DGL sparse library. Moreover, it achieves an average speedup of $2.16\times$ in end-to-end training compared to the popular GNN computing framework DGL. We open-source \name{} at \url{https://github.com/paoxiaode/DF-GNN}.
\end{abstract}

\maketitle

\section{Introduction}
Graph Neural Networks (GNNs) represent a category of deep learning techniques applied on graph-structured data, playing a vital role in diverse applications. Inspired by the success of the Transformer architecture~\cite{vaswani2017attention} in LLMs, Attention GNNs (AT-GNNs) have emerged as an important family of Graph Neural Networks. Various AT-GNNs models have been developed, including GAT family~\cite{velickovic2017graph, brody2021attentive}, Graph Transformer (GT)~\cite{gt,ying2021transformers,rampavsek2022recipe}, and AGNN~\cite{thekumparampil2018attention}. These models have demonstrated enhanced performance for node classification~\cite{kipf2016semi, xu2018powerful}, link prediction~\cite{zhang2018link,zhu2021neural}, and graph prediction tasks~\cite{errica2019fair, sun2019infograph}.

AT-GNNs are characterized by their complex computation pattern, typically encompassing three primary steps in each layer: firstly, calculating edge-wise attention scores using node and edge features; secondly, normalizing these scores across each node's local neighborhood; and finally, gathering and aggregating neighbor features, weighted by these normalized attention scores. To facilitate these operations, established GNN systems such as DGL~\cite{wang2019deep} and PyG~\cite{fey2019fast} implement the computation using the message passing paradigm~\cite{gilmer2017neural}, by breaking down GNNs into a series of independent kernels to support various model variants. Nonetheless, executing AT-GNN kernels sequentially on the GPU incurs considerable global memory access and CPU overhead from kernel launches, resulting in inefficient training of AT-GNN models, especially graph transformers, which may require several days to complete. To tackle these issues, several kernel fusion techniques have been proposed for general GNNs. Seastar~\cite{wu2021seastar} introduced a CUDA template for fusing multiple sparse operations, thereby omitting the materialization of edge data in GPU global memory. dgNN~\cite{zhang2022understanding} suggested a unified thread mapping scheme for both vertex and edge-centric operations to facilitate fusion and IO reduction. TLPGNN~\cite{fu2022tlpgnn} developed a two-level parallelism paradigm to circumvent atomic operations and enable coalesced memory access. However, applying these techniques to AT-GNNs encounters two significant challenges:

\begin{itemize}
\item \textbf{Varying compute patterns.} The computational process of AT-GNNs is marked by multiple operations of vertex-centric and edge-centric computations. This frequent change in computational patterns poses a significant challenge for existing kernel fusion strategies, which predominantly rely on uniform thread scheduling across all fused operations. Such a rigid scheduling framework struggles to adapt to the varying computational demands, often leading to inefficiencies. Furthermore, the inflexibility of fixed thread scheduling can result in unnecessary and redundant computations when the feature dimension diminishes in operations such as Softmax.
% \item \textbf{Shared memory bottleneck.} Prior solutions often leverage GPU shared memory to store intermediate data, minimizing data movement between GPU SMs and global memory. Fusing operations in AT-GNNs poses a need to store the intermediate attention weights in shared memory whose size grows proportional to the number of neighbors. This becomes particularly problematic in the presence of 'super nodes', which have high connectivity and can quickly deplete the available shared memory, limiting the performance gain from kernel fusion.
\item \textbf{Super node presence.} The computational performance of AT-GNNs is also influenced by the input graph. In real-world graphs, super nodes often exist with a significantly higher number of neighbors compared to other nodes in the same graph. The existence of super nodes leads to a performance bottleneck shift and can quickly deplete the available shared memory, limiting the performance gain from kernel fusion.
\end{itemize}

To address these challenges, we introduce \textbf{\name{}}, a framework designed for efficient GPU kernel fusion of AT-GNNs. The key to \name{} is \textit{allowing each operation in AT-GNNs to dynamically adopt the optimal thread scheduling strategy, while also enjoying the benefits of kernel fusion, including shared memory usage and reduced kernel launch overhead}. 
\textbf{At the GPU kernel scheduling level}, \name{} incorporates a Dynamic Bi-level Thread Scheduling (BTS) design. Dynamic BTS allows each operation to explore its optimal thread scheduling strategy between and within thread blocks. For intra-block scheduling, \name{} addresses the specific performance bottlenecks of the three operations in AT-GNNs and devises tailored scheduling strategies. In inter-block scheduling, \name{} accounts for performance bottlenecks introduced by super nodes and formulates two general kernel fusion methods, selecting the appropriate method based on the input graph's characteristics.
\textbf{At the model level}, \name{} extends optimization to the backward pass, utilizing the same fusion strategy to optimize the backward computation, enabling acceleration in both forward and backward computations during AT-GNNs training. 
\textbf{At the framework level}, \name{} is integrated with the PyTorch~\cite{paszke2019pytorch} framework and provides a series of APIs for easy user access. Users can easily invoke the \name{} APIs within their PyTorch models to facilitate efficient model training and inference.

Through comprehensive evaluations on multiple AT-GNNs models and diverse datasets, \name{} consistently outperforms existing GNN optimization works such as cuGraph and dgNN, with speedups over the DGL sparse baseline ranging from $1.92\times$ to $7.00\times$. Additionally, \name{} achieves an average end-to-end training speedup of $2.16\times$ compared to the DGL framework. These results underscore the effectiveness of \name{} in enhancing both kernel-level and end-to-end performance for AT-GNNs training.

\section{Background}
\label{sec:background}

\subsection{Attention GNN Vectorized Formulation}

The message passing paradigm presents the computation of GNN from the view of an individual node and its local neighbors. However, this node-centric view can't efficiently capture operations on a graph which consists of a batch of nodes and edges. Thus, a vectorized formulation of AT-GNNs is desired. DGL~\cite{wang2019deep} proposed to generalize classical operations for sparse linear algebra as the vectorized abstraction of the message passing paradigm. Specifically, one generalized Sampled Dense-Dense Matrix Multiplication (SDDMM) is responsible for generating the message on edges, while a generalized Sparse-dense Matrix Multiplication (SpMM) is in charge of message aggregation. Applying the idea to AT-GNNs gives the following formulation:
\begin{equation}
    \begin{aligned}
    \mathbf{Q}^{N\times d}, \mathbf{K}^{N\times d}, \mathbf{V}^{N\times d} &= f_Q(\mathbf{X}), f_K(\mathbf{X}), f_V(\mathbf{X})\\
    \text{Message:   } \mathbf{S}^{N\times N} &= \text{SDDMM}(\mathbf{Q}, \mathbf{K},\mathbf{A})\\
    &= (\mathbf{Q} *_{\otimes\oplus} {\mathbf{K}}^\top) \odot \mathbf{A}, \\
    \text{Normalize:  } \mathbf{P}^{N\times N} &= \text{Softmax}(\mathbf{S}),  \\
    \text{Aggregate:  } \mathbf{O}^{N\times d} &= \text{SpMM}(\mathbf{P}, \mathbf{V}) \\
    &= \mathbf{P} *_{\otimes\oplus} \mathbf{V}.\\
\end{aligned}
\label{equ:graph_conv}
\end{equation}
Here, $N$ is the number of nodes in graph $G$, $d$ is the node feature dimension, $\mathbf{A}$ represents the adjacency matrix, and $\mathbf{X}$ represents the node feature matrix. The message step first transforms node features $\mathbf{X}$ into $\mathbf{Q}$, $\mathbf{K}$ and $\mathbf{V}$ matrices and then invokes an SDDMM operation to compute attention weights $\mathbf{S}$. $*_{\otimes\oplus}$ is a generalized matrix multiplication equipped with a scalar addition and a scalar multiplication, while $\odot$ is the Hadamard product (or element-wise multiplication). The flexibility in the choice of $\oplus$ and  $\otimes$ leads to different graph convolution variants in computing attention and aggregating messages.

% Another notable complexity in the vectorized formulation of AT-GNNs is the Softmax operation, which normalizes the sparse attention weight matrix $\mathbf{S}$. To enhance the numeric stability of Softmax, its implementation adopts a normalization trick as shown in Equation ~\eqref{equ:softmax} that involves two rounds of information exchange between nodes and edges: first one to obtain the maximum value $z^{(max)}v$ of its residing row, and a second one to sum up all connected edge values $f_{uv}$.

% \begin{equation}
% \begin{aligned}
% \text{Round.1: } z^{(max)}_v &= \max_{u\in \mathcal{N}(v)}(s_{uv}) \\
% f_{uv} &= \exp(m_{uv} - z^{(max)}_v) \\
% \text{Round.2: } z^{(sum)}_v &=\sum_{u\in \mathcal{N}(v)} f_{uv} \\
% p_{uv} &= \frac{f_{uv}}{z^{(sum)}_v} \\
% \end{aligned}
% \label{equ:softmax}
% \end{equation}

\subsection{Existing Optimizations and Limitations}

% \stitle{Bi-level Thread Scheduling (BTS).} In GPUs, a kernel is executed by a group of parallel thread blocks, each containing up to 1024 threads. Therefore, the GPU kernel thread scheduling strategy can be categorized into two levels: \textit{\textbf{inter-block thread scheduling}} and \textit{\textbf{intra-block thread scheduling}}. Inter-block thread scheduling distributes workload across different thread blocks for parallel execution. Intra-block thread scheduling manages threads within the same block to execute specific operations on divided workloads.

\begin{figure}[t]
    \centering
    \begin{minipage}[b]{.45\textwidth}
        \centering
        \includegraphics[width=0.9\textwidth]{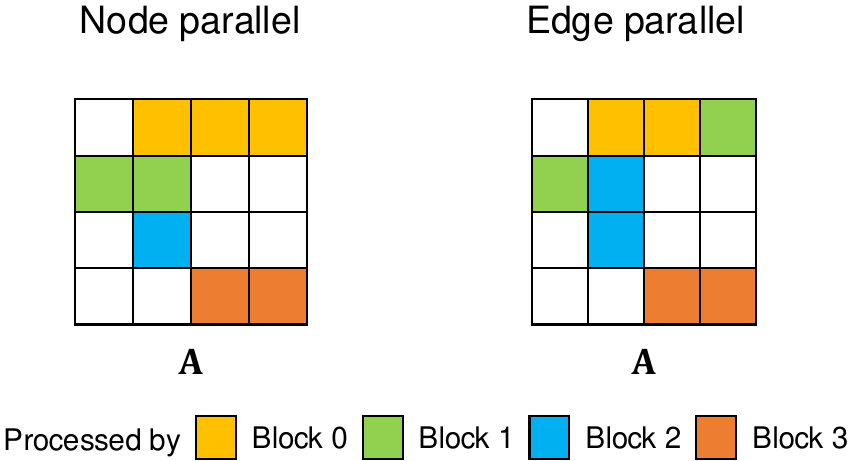}
        \caption{Workload partition in node and edge parallel.}
        \label{fig:node-edge_para}
    \end{minipage}%
    \hspace{0.05\textwidth}
    \begin{minipage}[b]{.45\textwidth}
        \centering
        \includegraphics[width=0.9\textwidth]{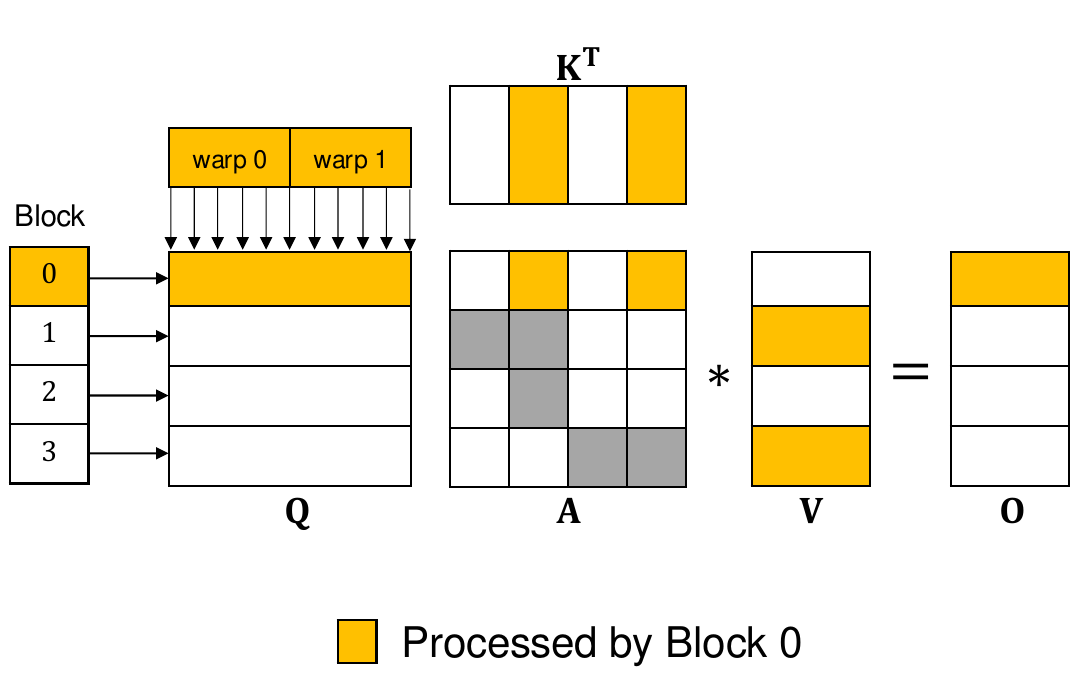}
        \caption{The fused GT convolution kernel under feature parallel fusion strategy.}
        \label{fig:row-wise-fusion}
    \end{minipage}
\end{figure}

\stitle{Parallel strategies for GNN operations.}  In GNN-related workload, inter-block scheduling have two typical strategies: \textit{\textbf{Node parallel}}~\cite{wang2019deep}, which allocates the workload of a node and its neighbors to one parallel unit, naturally fits the computational pattern of SpMM which aggregates values according to nodes. \textit{\textbf{Edge parallel}}~\cite{wang2019deep}, in contrast, assigns different parallel units to different edges, achieving ideal workload balance for SDDMM. Figure \ref{fig:node-edge_para} depicts the difference between node and edge parallel. In addition to these two inter-block methods, \textit{\textbf{Feature parallel}}~\cite{fu2022tlpgnn} serves as an intra-block scheduling method. Within the thread block, the number of threads corresponds to the feature dimension, with each dimension mapped to a thread. Based on these parallel strategies, various approaches have been proposed to accelerate individual operation kernels such as SpMM or SDDMM. However, without kernel fusion, AT-GNNs involving multiple operations will invoke a series of fragmented kernels. This fragmentation can lead to performance loss, as these kernels fail to efficiently utilize shared memory to minimize data movement and often incur significant CPU overhead from frequent kernel launches.

% \stitle{Kernel fusion in GNN.} Motivated by the challenges brought by fragmented kernels, another category of approaches~\cite{zhang2022understanding,rahman2021fusedmm,fu2022tlpgnn,xie2022graphiler} strives to fuse graph convolution, which contains SpMM, Softmax and SDDMM operations, into a single kernel. The kernel fusion aims to reduce communication between Streaming Multiprocessors (SMs) and global memory, overlap data retrieval and computation, and mitigate kernel launch overhead. ~\cite{zhang2022understanding} further proposes to re-order certain operations to eliminate redundant computation caused by message propagation. 

\stitle{Kernel fusion.} Kernel fusion is a widely used optimization technique that effectively addresses the challenges posed by fragmented kernels, and a series of works represented by FlashAttention~\cite{dao2022flashattention} have achieved great success in LLMs field. In GNN, research on kernel fusion can be categorized into two primary approaches: one is handcrafted fused kernels~\cite{zhang2022understanding,rahman2021fusedmm,fu2022tlpgnn}, manually writing optimized CUDA kernels. These optimizations typically target specific, fixed computation patterns found in certain GNN convolutions. The other is compiler-based frameworks~\cite{xie2022graphiler,wu2021seastar,hu2020featgraph}, which automatically generate fused kernels based on user-defined functions. However, whether through manual coding or automated generation, existing approaches in GNN field generally employ a fixed parallel strategy for all operations within the fused kernel. This includes inter-block scheduling utilizing node parallel and intra-block scheduling employing feature parallel, as Figure \ref{fig:row-wise-fusion} shows. In the following text, we refer to this fixed fusion approach as \textit{\textbf{feature parallel fusion strategy}}.

The feature parallel fusion strategy encounters two primary hurdles when employed in the context of AT-GNNs, stemming from the irregularity in computation and input data. The first challenge revolves around computation irregularity, where the three operations within AT-GNNs have different optimal parallel strategies. Specifically, SDDMM is suited for the edge parallel strategy, while Softmax and SpMM are better suited for the node parallel strategy. The feature parallel fusion strategy fails to account for these differences.
The second challenge arises from the irregularity of input graph. Real-world graphs have diverse characteristics, and some contain super nodes. These variations can shift the performance bottleneck of graph convolution, preventing kernel fusion from achieving optimal performance in certain cases. However, the feature parallel fusion strategy does not consider the data characteristics of the input graphs. 
\section{Accelerate AT-GNNs with \name{}}
\label{sec:framework}

\begin{figure}[t]
    \centering
    \begin{minipage}[t]{.45\textwidth}
        \centering
        \includegraphics[width=\linewidth]{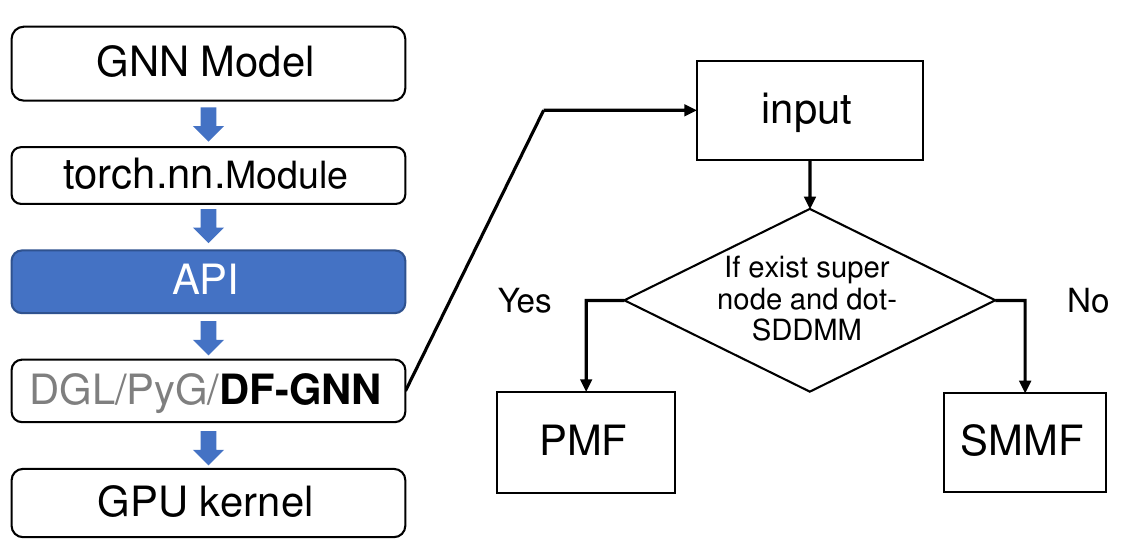}
        \caption{\name{} framework.}
        \label{fig:DF-GNN-framework}
    \end{minipage}
    \hspace{0.05\textwidth}
    \begin{minipage}[t]{.45\textwidth}
         \centering
        \includegraphics[width=\linewidth]{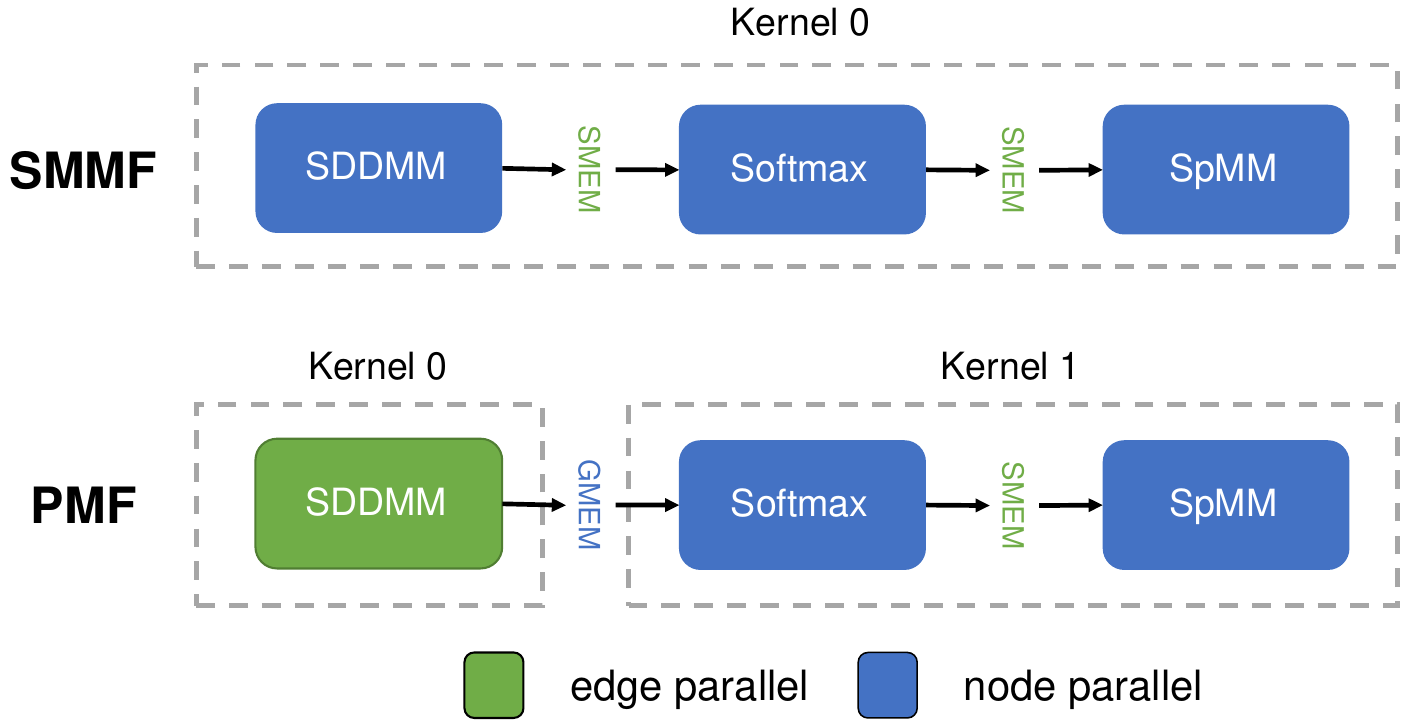}
        \caption{The \allfusedabb{} and \parfusedabb{} in \name{}.}
        \label{fig:dynamic_thread_schedule}
    \end{minipage}%

\end{figure}

% \name{} implements two general kernel fusion methods for AT-GNNs: \allfused{} \textit{\textbf{(\allfusedabb{})}} and \parfused{} \textit{\textbf{(\parfusedabb{})}}, which maximizes the utilization of shared memory and parallelism respectively. Figure \ref{fig:dynamic_thread_schedule} shows the inter-block scheduling in the \allfusedabb{} and \parfusedabb{}. For the \allfusedabb{}, three operations share the same node parallel inter-block scheduling and are fused into a single kernel, with intermediate variables passed between operations through shared memory. Meanwhile, at the intra-block scheduling level, warp-balanced SDDMM, redundancy-free Softmax and vectorized SpMM are used for optimization. For the \parfusedabb{}, SDDMM employs edge parallel while Softmax and SpMM employ node parallel. Softmax and SpMM are fused, with SDDMM passing its result to them through global memory. Additionally, the \parfusedabb{} also incorporates redundancy-free Softmax and vectorized SpMM.

\name{} framework is developed to be fully compatible with the widely-used deep learning framework PyTorch, similar to existing GNN frameworks like DGL and PyG, as illustrated in Figure~\ref{fig:DF-GNN-framework}. Users can easily invoke the optimized fused GPU kernel code through the \name{} graph convolution API within the \textit{torch.nn.Module}, thereby enhancing the efficiency of GNN models.

The framework supports two general kernel fusion methods for AT-GNNs: \allfused{} \textit{\textbf{(\allfusedabb{})}} and \parfused{} \textit{\textbf{(\parfusedabb{})}}, which maximizes the utilization of shared memory and parallelism, respectively, as depicted in Figure \ref{fig:dynamic_thread_schedule}. For the \allfusedabb{}, three operations are fused into a single kernel, sharing the same node parallel inter-block scheduling, with intermediate variables transferred through shared memory. Conversely, the \parfusedabb{} employs edge parallel SDDMM while utilizing node parallel for Softmax and SpMM operations. In this case, Softmax and SpMM are fused, with SDDMM passing its results via global memory. Additionally, \name{} enables dynamic selection between \allfusedabb{} and \parfusedabb{} at runtime with each API call, with the rationale for this choice discussed in detail in Section \ref{sec:intra thread schedule}.

\section{\name{} Design}
\label{sec:design}

\stitle{Overview.} \name{} employs a dynamic bi-level thread scheduling method, enabling operations to adopt their optimal thread scheduling strategies and leverage shared memory for efficient inter-operation communication within a fused kernel. Based on this method, \name{} designs tailored thread scheduling strategies for three operations in AT-GNNs, which include warp-balanced SDDMM, redundancy-free Softmax, and vectorized SpMM. Moreover, \name{} also fuses operations in backward computation to accelerate end-to-end AT-GNNs training. 

\subsection{Dynamic Bi-level Thread Scheduling}
To address the challenges of AT-GNNs kernel fusion, \name{} designs \textit{\textbf{Dynamic Bi-level Thread Scheduling}}, a fine-grained, dynamic kernel thread scheduling strategy to replace the previous coarse-grained, fixed scheduling schemes, allowing the three operations in AT-GNNs to employ different parallel strategies. For the inter-block scheduling, operations can select between edge parallel and node parallel strategies, based on factors such as the need to avoid atomic operators and the patterns of graph convolution computation. For the intra-block scheduling, operations may apply different strategies based on their own computational characteristics. If an operation uses a different intra-block scheduling than the preceding operation, \name{} enforces thread synchronization within the thread blocks and reschedules the threads according to the new scheduling. Additionally, \name{} fuses adjacent operations that share the same inter-block scheduling into a single kernel, leveraging shared memory for efficient operation communication within the fused kernel. 

% 相较于fixed strategy的优势
The dynamic thread scheduling design offers two primary advantages over the fixed feature parallel fusion strategy in FusedMM~\cite{rahman2021fusedmm}, TLPGNN~\cite{fu2022tlpgnn}, and dgNN~\cite{zhang2022understanding}.
% 拓展了kernel scheduling的优化空间
Firstly, the bi-level flexible thread scheduling between and within thread blocks expands the optimization space for kernel performance. Each operation can choose optimal scheduling strategies at both levels based on the compute pattern, enhancing the performance gain of each operation within the \name{} kernels.
% 提高了fusion strategy的全面性
Secondly, the flexible combination of different scheduling strategies empowers \name{} to consider input data factors when deciding the thread scheduling strategy. For specific input graph characteristics, \name{} adopts different kernel scheduling strategies to address their unique bottlenecks, improving the generality of \name{}'s fusion strategy. 
% % \mj{The second point is the same as the first point}
% % 降低了smem的使用门槛
% Thirdly, the kernel fusion of adjacent operations with the same inter-block scheduling enables the use of GPU shared memory. The output data can be stored in shared memory and directly consumed by the next operation, thereby minimizing data movement between GPU streaming multiprocessors and global memory.

\begin{figure*}[t]
    \centering
    \includegraphics[width=\textwidth]{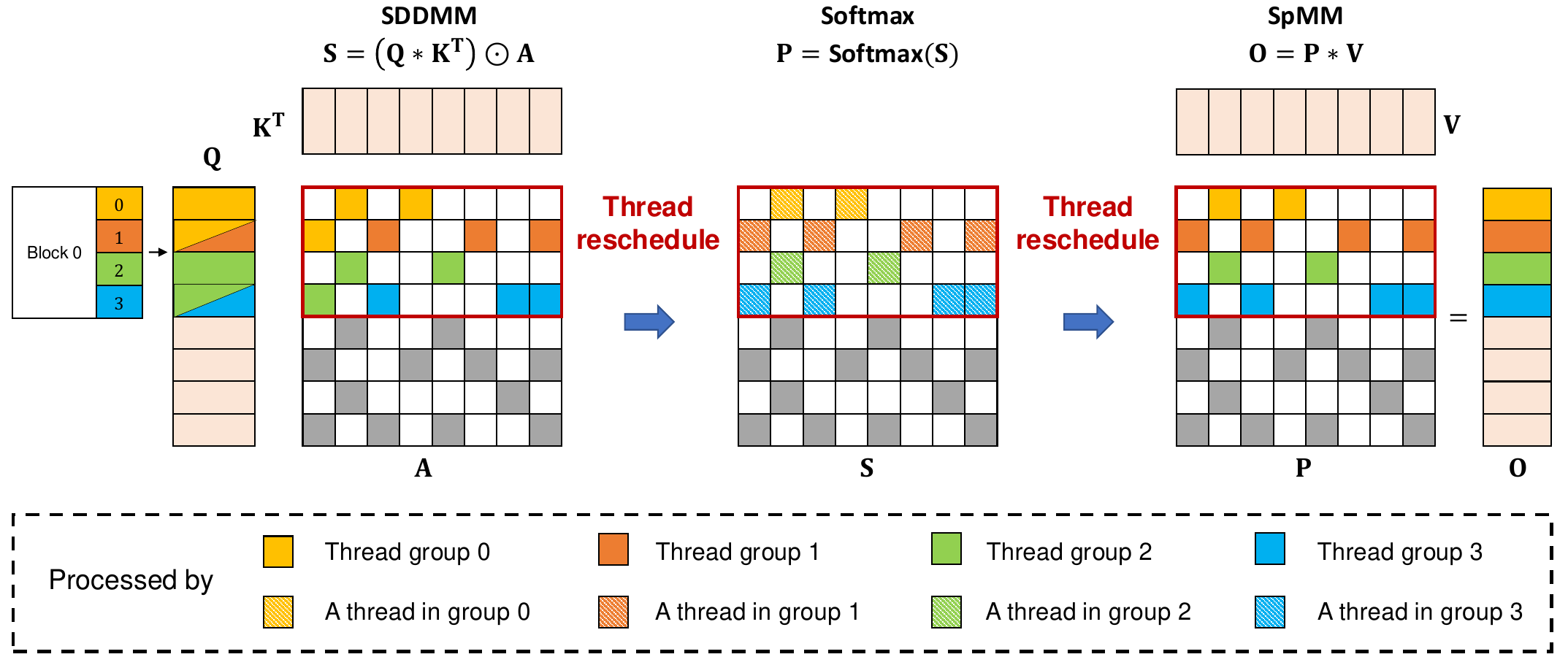}
    % \caption{GT convolution (SDDMM-Softmax-SpMM) in \name{}.}
    \caption{Dynamic Bi-level Thread Scheduling in GT convolution.}
    \label{fig:Balanced-fusion}
\end{figure*} 

Figure~\ref{fig:Balanced-fusion} illustrates the application of dynamic BTS within the graph transformer model. In this instance, the \allfusedabb{} approach is utilized, where three operations are fused into a single kernel that shares the same node parallel inter-block scheduling. Each thread block comprises four thread groups (warps) and is responsible for managing the computations of four rows. Meanwhile, the intra-block scheduling is dynamically adjusted for different operations to optimize kernel performance. In SDDMM, the edges across these four rows are evenly distributed among the four thread groups for processing. Following this, thread groups adjust their scheduling to node parallel for both Softmax and SpMM since both of them require row-wise aggregations. Notably, since Softmax does not involve feature dimension, threads within a thread group are flattened to process a single row. While in SpMM, a thread group collectively processes an edge, iterating over the corresponding row.

\subsection{Tailored Thread Scheduling}
\label{sec:intra thread schedule}

% The dynamic BTS provides a flexible thread scheduling mechanism at the intra-block level, offering more optimization opportunities. Based on this, \name{} devises tailored scheduling strategies for the unique bottlenecks of each operation. SDDMM employs two thread scheduling approaches: warp-balanced and edge parallel, chosen based on the specific characteristics of the graph. For Softmax and SpMM, \name{} introduces redundancy-free Softmax and vectorized SpMM techniques, eliminating redundant computation and reducing memory access overhead.

\begin{figure}[t]
    \centering
    \begin{minipage}[t]{.45\textwidth}
            \centering
            \includegraphics[width=\linewidth]{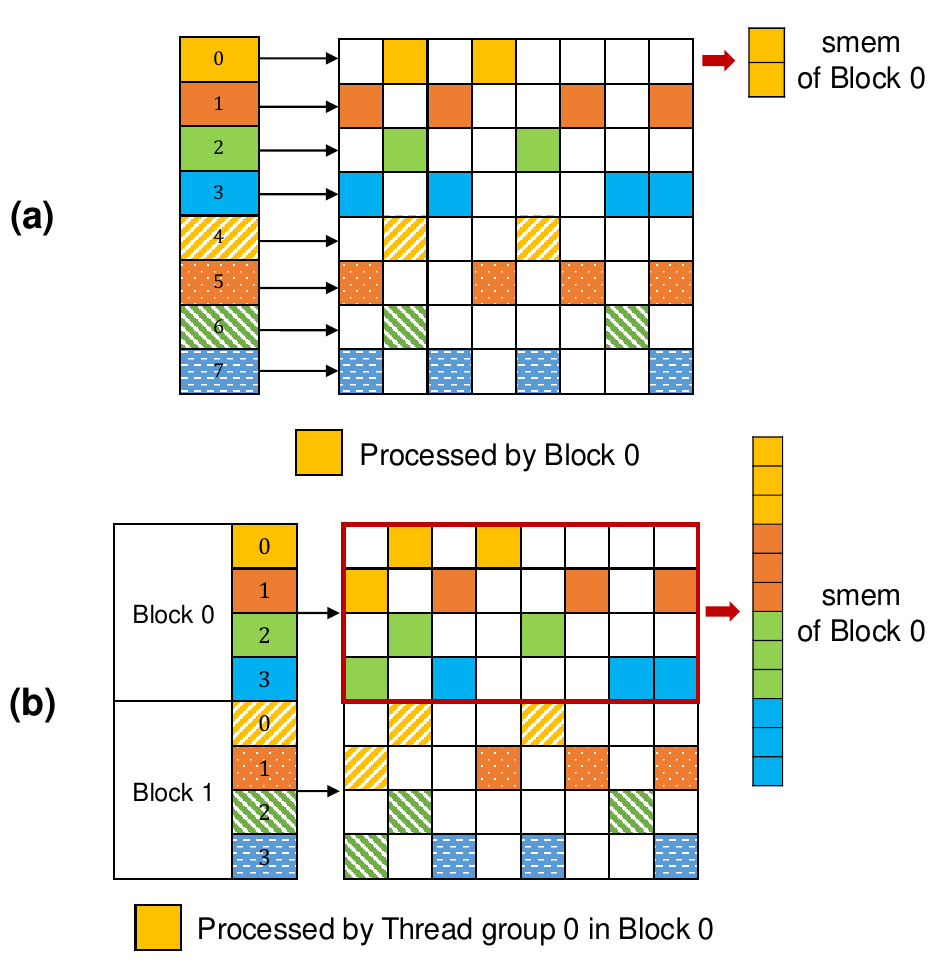}
            \caption{(a) Feature parallel SDDMM. (b) Warp-balanced SDDMM in \name{}.}
            \label{fig:sddmm-balanced}
    \end{minipage}%
    \hspace{0.05\textwidth}
    \begin{minipage}[t]{.45\textwidth}
            \centering
            \includegraphics[width=\linewidth]{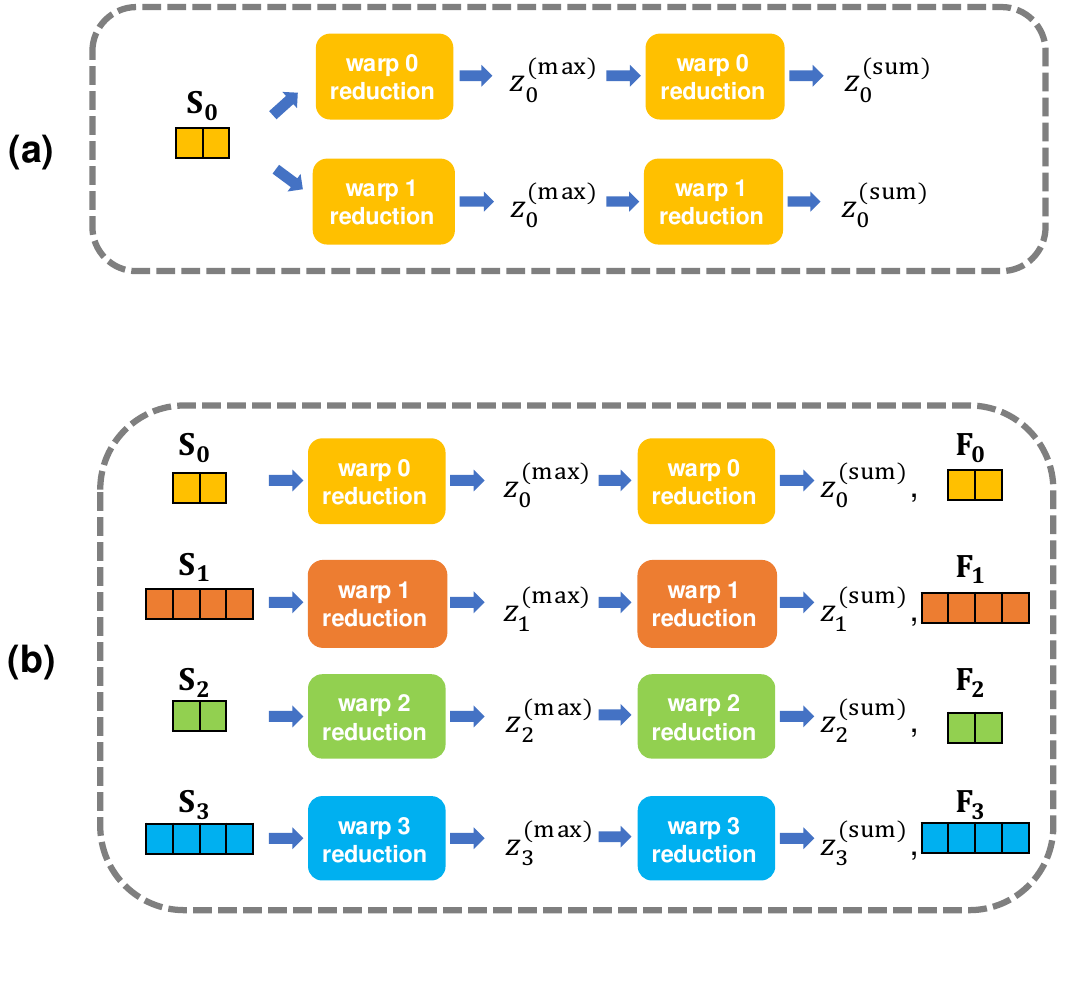}
            \caption{(a) Feature parallel Softmax. (b) Redundancy-free Softmax in \name{}.}
            \label{fig:softmax-all}
    \end{minipage}
\end{figure}

\stitle{SDDMM.} In the existing inter-block thread scheduling strategies, edge parallel evenly distributes edges across all thread blocks, making it most suitable for SDDMM where each edge result is independent. However, edge parallel SDDMM is difficult to fuse with the subsequent Softmax and SpMM operations, which are better suited for node parallel. Consider this, node parallel SDDMM is used in existing feature parallel fusion strategy, which processes edges in one row by a thread block, as shown in Figure~\hyperref[fig:sddmm-balanced]{6a}. Nevertheless, node parallel SDDMM suffers from workload imbalance between thread blocks due to the varying numbers of neighbors.

Inspired by the workload balancing works in \cite{ye2023sparsetir,yu2021exploiting}, \name{} addresses the issue of SDDMM workload imbalance and facilitates kernel fusion by designing the \textit{\textbf{warp-balanced SDDMM}}, adopting a balanced strategy with intra-block scheduling using edge parallel and inter-block scheduling using node parallel. Figure~\hyperref[fig:sddmm-balanced]{6b} illustrates the process of warp-balanced SDDMM. In this approach, threads within a thread block are grouped, and these thread groups evenly distribute edges and loop over them to compute the SDDMM result. The computed attention weights of a thread block are then stored in shared memory, ready for subsequent utilization in the Softmax and SpMM operations. Compared to the feature parallel strategy, the tailored SDDMM thread scheduling in \name{} ensures warp-level workload balance. Warp-balanced SDDMM uniformly distributes computations for multiple adjacent nodes and enables kernel fusion with subsequent operations, resulting in a significant reduction in overall imbalance and kernel execution time.

% \begin{equation}
% \begin{aligned}
%         \max_v(\mathcal{N}(v)) &\geq \frac{\text{Shared memory capacity}}{\text{Bytes of feature data type}}\\
% \end{aligned}
% \label{equ:super_node}
% \end{equation}

Warp-balanced SDDMM employs intra-block edge parallel outperforms existing feature parallel SDDMM. However, the performance of the fused kernel is also influenced by the input graph data and the compute pattern of graph convolution. In some cases, edge parallel SDDMM may be more effective than node parallel warp-balanced SDDMM. The cost of the SDDMM operation can be expressed as \textit{$\text{cost}_{\text{SDDMM}}=$ the number of edges $\times$ the cost of the operator per edge}. This implies that the cost for SDDMM increases in two scenarios: first, when the input graph is large, especially if it contains super nodes with numerous neighbors, making it impossible to store SDDMM results in shared memory; and second, when a high-cost edge operator, such as dot, is used in SDDMM. Considering this, if a graph contains super nodes that satisfy the condition $\max_v(\mathcal{N}(v)) \geq \frac{\text{Shared memory capacity}}{\text{Bytes of feature data type}}$ and employs dot-SDDMM, we propose that the performance bottleneck of the fused kernel shifts from CPU launch overhead to unbalanced and low-utilization SDDMM. In these scenarios, \name{} utilizes \parfusedabb{}, which employs edge-parallel SDDMM. Conversely, if the condition is not met, \allfusedabb{} is adopted, incorporating warp-balanced SDDMM, as illustrated in Figure~\ref{fig:DF-GNN-framework}.

\stitle{Softmax.} Integrating the Softmax operation into a fused kernel poses a significant challenge due to the absence of the feature dimension. The existing feature parallel fusion strategy, which aligns the number of threads with the feature dimension, is suitable for operations like SDDMM and SpMM that process along this dimension. However, when the feature dimension is absent in the Softmax step, feature dimension alignment introduces duplicated computations. As depicted in Figure~\hyperref[fig:softmax-all]{7a}, two warps in the same block have to execute identical operations in parallel, including reading $S_v$ from shared memory, performing a max warp reduction, and conducting a sum warp reduction. Additionally, feature parallel fusion fails to cache the reusable Softmax result $\mathbf{F_v}$, resulting in its recomputation during the subsequent SpMM step.

To solve the above challenge in Softmax, \name{} designs \textit{\textbf{redundancy-free Softmax}}, rescheduling threads along the edge dimension rather than the feature dimension to avoid duplicated computations. In this approach, each thread group manages the computation of a row, where threads in a group are flattened to handle the corresponding edges. Warp-level reduction is used to aggregate the attention weights for normalization.
The computation of the redundancy-free Softmax is illustrated in Figure~\hyperref[fig:softmax-all]{7b}. For instance, warp 0 is responsible for processing $\mathbf{S}_0$ of row 0, while warp 1 is assigned to handle $\mathbf{S}_1$ of row 1. Distinct warps do not execute identical tasks, enabling the efficient utilization of hardware resources. Additionally, during the second warp reduction, the redundancy-free Softmax saves the intermediate values $\mathbf{F}_v$ into shared memory. This enables the reuse of these values in the subsequent SpMM step, avoiding the need for recomputation.

% \begin{figure}[t]
%     \centering
%     \includegraphics[width=0.45\textwidth]{figure/vectorized_spmm.pdf}
%     \caption{Vectorized SpMM in \name{}}
%     \label{fig:vectorized_spmm}
% \end{figure}

\stitle{SpMM.} The SpMM operation, which aggregates the weighted node features to output features, is a key component of the fused AT-GNNs kernel. SpMM operation incorporates with a large number of accesses to global memory, making memory the primary bottleneck affecting SpMM performance.

To address the memory bottleneck, \name{} implements \textit{\textbf{vectorized SpMM}} which incorporates \textit{vectorized memory access} and \textit{shared memory cache}. Due to the mismatch between the number of threads and the feature dimension, SpMM involves two nested loops: one loop iterates over the feature dimension, and the other iterates over the neighbors. In \name{}, the outer loop processes the feature dimensions using vectorized memory access instructions, while the inner loop iterates over neighbors to accumulate intermediate results in shared memory. 
This shared memory cache reduces the need for repeated global memory accesses. Furthermore, employing vectorized memory access decreases the number of memory access instructions in SpMM. These optimizations significantly lower the memory overhead of SpMM and improve the performance of the fused kernel.

\subsection{Backward Pass Kernel Fusion Optimization}

\begin{figure}[t]
\centering
\includegraphics[width=0.8\linewidth]{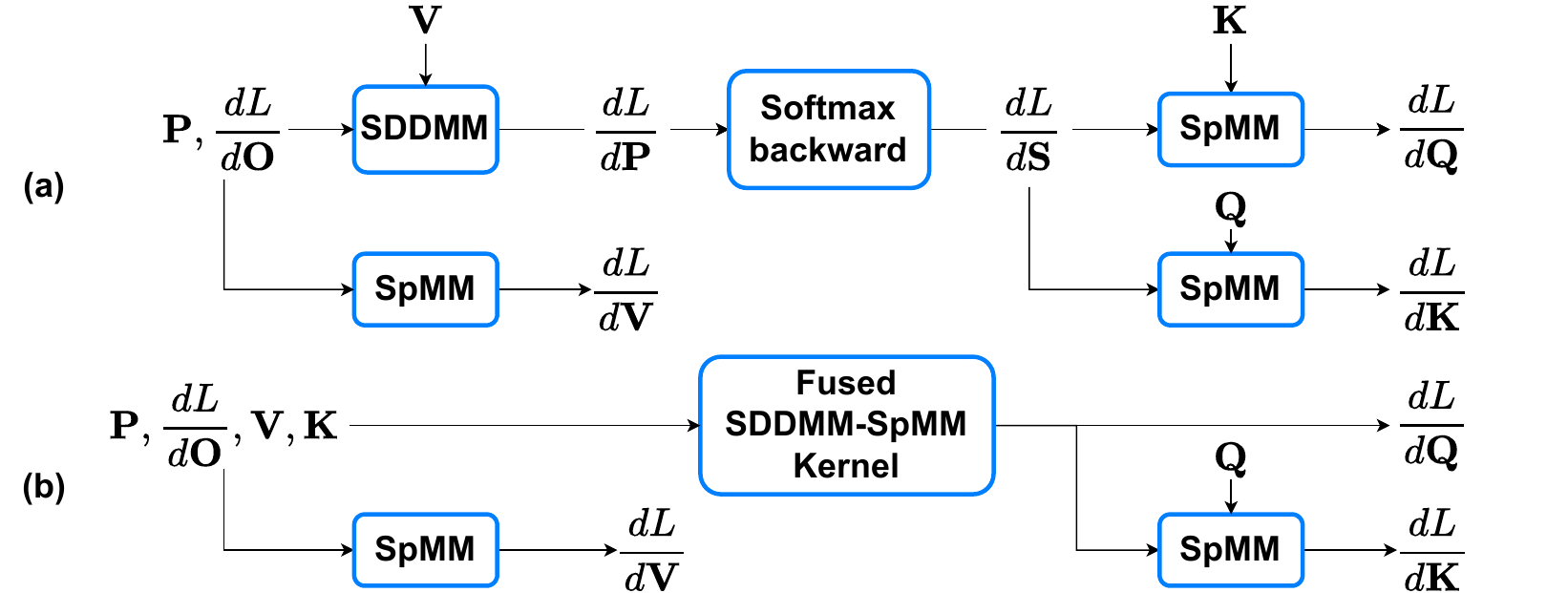}
\caption{ Kernels in GT backward pass. (a) Without any fusion. (b) After \name{} fusion. }
\label{fig:Q_bw}
\end{figure}
In \name{}, we extend the kernel fusion optimization to the backward pass.
This extension is motivated by the observation that \textit{the backward pass of SDDMM and SpMM also involve SDDMM and SpMM operations}~\cite{wang2019deep}. The detailed formulations for these backward operations can be found in the appendix~\ref{sec:bwd_at_gnn}. Figure~\ref{fig:Q_bw} provides an overview of the GT backward computation. The computation of $\frac{\mathrm{d}L}{\mathrm{d}\mathbf{Q}}$ consists of three steps: SDDMM, Softmax backward and SpMM, mirroring the SDDMM, Softmax and SpMM pattern in the forward pass. By leveraging this similarity, \name{} applies the same fusion strategy to fuse these three kernels into a single fused SDDMM-SpMM kernel. This unified approach eliminates the need for a distinct fusion strategy in the backward pass, reducing the number of required kernels in backward and significantly accelerating the model training.

\section{Experiment Evaluation}

% \begin{figure*}[t]
%     \centering
%     \includegraphics[width=\textwidth]{figure/batch_graph_kernel_perf_f128_new.pdf}
%     \caption{Normalized kernel time speedup and bandwidth utilization for three GNN models.}
%     \label{fig:kernel_bs1024_k64}
% \end{figure*}

% In this Section, we conduct extensive experiments to evaluate the performance of \name{}. We first introduce the experiment settings in Section~\ref{exp:settings}. Next, we compare \name{} with state-of-the-art GNN systems in Section~\ref{exp:main} on three types of graph datasets. Then, we evaluate the end-to-end speedup of \name{} in Section~\ref{exp:endtoend}, and finally use ablation experiments to evaluate the performance of \name{} under individual design decisions and different input data settings in Section~\ref{exp:ablation}.

\subsection{Experiment Setup}\label{exp:settings}

\begin{table}[h]
\vspace{-3mm}
\caption{Datasets for evaluation, sorted by avg. degree, $^*$ indicates super nodes exist.}
\centering
\scriptsize
\label{dataset_eval}
\begin{tabular}{l|lcccc}
\hline
Type & Dataset&Abbr.  & \#Node & \#Edge & avg/max degree \\ \hline
    \multirow{6}{*}{\thead{Full\\graph}}  &  Citeseer&CS  &  3.3K  &  9.2K  &    2.8 / 99       \\
 &  Cora&CR  &  2.7K  &  10K &   3.9 / 168         \\
 &  Pubmed &PB  &  19K&  88K  &   4.5 / 171       \\
  % &  Arxiv & AR  &  169K&  1.1M  &   6.9 / 436       \\
 % &  Flickr & FL  &  89K  &  899K  &    10 / 5425       \\
 % &  Yelp & YP  &  716K  &  13M  &    19.5 / 4886       \\\
 &  $\text{Ogbg-ppa}^*$ & PP  &  576K  &  42M  &    74 / 3241       \\
 &  $\text{Reddit}^*$ & RD  &  232K &  114M &    492 / 21657       \\
 &  $\text{Protein}^*$ & PR  &  132K  &  79M  &   597 / 7750         \\ \hline 
% \multirow{7}{*}{\thead{Batch\\graph}}   &  Peptides&PT  &  151  &  307  &     2    \\ 
%  &  COCO-SP&CO  &  477 &  2693    &   3.9     \\ 
 \multirow{6}{*}{\thead{Batch\\graph}}   &  COCO-SP&CO  &  477 &  2693    &   5.6    \\ 
 &  PascalVOC &PV  &  479  &  2709  &   5.6        \\ 
 % &  PascalVOC-SP &PV  &  479  &  2709  &   5.6        \\ 
 &  MNIST & MN   &  71  &   564     &   8      \\
 &  CIFAR10 & CF  &  118  &  941   &   8      \\ 
 &  CLUSTER & CL  &  117  &  4303  &   36.7        \\
 &  PATTERN & PN   &  119  &  6079  &   51.1   \\ 
\hline
\end{tabular}
\vspace{-3mm}
\end{table}

\stitle{GNN models.} We choose three representative AT-GNNs with the SDDMM-Softmax-SpMM pattern. 1) \textit{\textbf{GT}}~\cite{gt} is similar to the NLP transformer architecture~\cite{vaswani2017attention}, but leverages a sparse attention mechanism by only computing the attention scores between the features of neighboring nodes with dot-SDDMM; 2) \textit{\textbf{AGNN}}~\cite{thekumparampil2018attention} is similar to GT, but with an $L_2$ normalization on input node features; 3) \textit{\textbf{GAT}}~\cite{velickovic2017graph} is the most classic AT-GNNs model with add-SDDMM, utilizing attention mechanisms to capture complex relationships within graph-structured data.

\stitle{Baselines.} We compare \name{} with the following state-of-the-art baselines in our experiments. 1) \textit{\textbf{DGL sparse}}~\cite{wang2019deep} is a sparse operator library under DGL framework and offers a variety of non-fused graph sparse operators, including SpMM and SDDMM; 2) \textit{\textbf{PyG}}~\cite{fey2019fast} is a popular GNN framework in which users can utilize message passing abstract to define graph convolutions; 3) \textit{\textbf{dgNN}}~\cite{zhang2022understanding} is a SOTA open-source research work that fuses multiple operators of a GNN models into a single GPU kernel, utilizing a feature parallel fusion strategy, as discussed in Section~\ref{sec:background}.  4) \textit{\textbf{cuGraph}}~\cite{cugraph} is an advanced software library developed by NVIDIA, providing APIs to access highly-optimized handcrafted GNN kernels.

% We use DGL sparse as a baseline without kernel fusion for comparison

\stitle{Datasets.} Table~\ref{dataset_eval} lists the datasets used for evaluation. Batch graph datasets contains numerous small graphs, which are normally used for graph classification tasks. We use the datasets commonly employed for AT-GNNs, including datasets from Benchmarking GNNs~\cite{dwivedi2023benchmarking} (PATTERN, CLUSTER, MNIST, CIFAR10) and Long-Range Graph Benchmark~\cite{dwivedi2022long} (PascalVOC-SP, COCO-SP). Running AT-GNNs on full graph can be used for node classification tasks. We choose three popular datasets (Cora, Citeseer, and Pubmed), along with three datasets with super nodes (Ogbg-ppa, Reddit, Protein).  

\stitle{Environments.}
The hardware we used for evaluation is an AWS g5.8xlarge server with one 32-core, 64 thread AMD EPYC 7R32 CPU @ 2.8GHz and a NVIDIA A10G GPU with CUDA 12.2. Regarding the software, we use PyTorch 2.3, torch geometric 2.5 and DGL 2.1. 
% Our kernels are implemented using the CUDA backend and integrated with PyTorch by its custom C++ extensions.

\stitle{Model Settings.}
The feature dimension and batch size are set to $128$ and $1024$. To ensure a fair comparison, \name{} kernels are configured to produce results consistent with the baselines.

\begin{figure*}[t]
    \centering
\includegraphics[width=\textwidth]{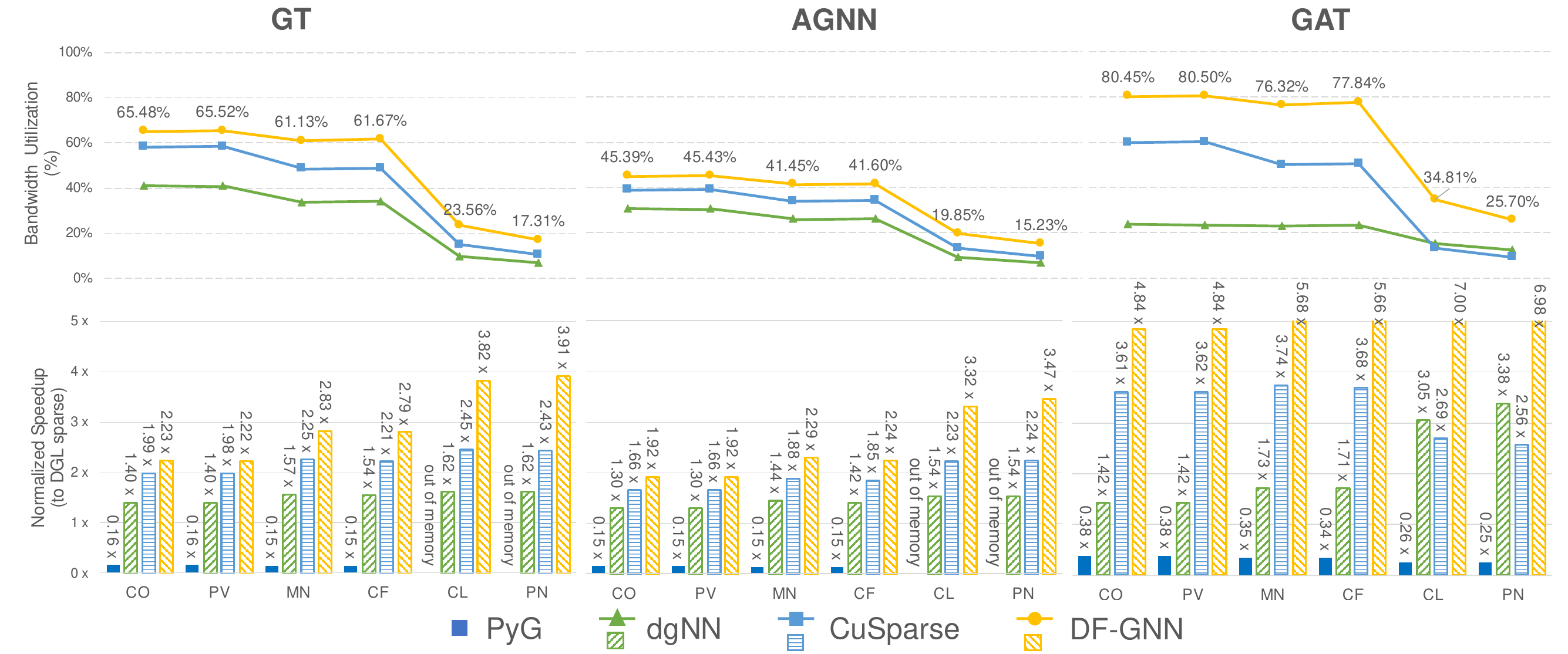}
    \caption{Normalized kernel time speedup and bandwidth utilization on batch graph datasets.}
    \label{fig:kernel_bs1024_k64}
\end{figure*}
\begin{figure*}[t]
    \centering
    \includegraphics[width=\textwidth]{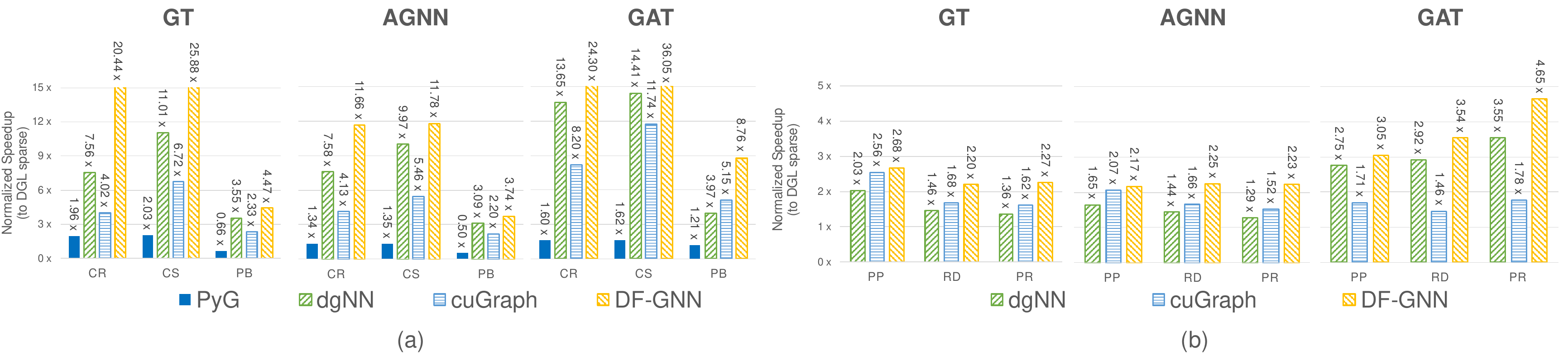}
    \caption{Normalized kernel time speedup on (a) Full graphs and (b) Full graphs with super nodes.}
    \label{fig:full-graph-perf}
\end{figure*}
\subsection{Kernel Performance}\label{exp:main}

\vspace{-3mm}
\stitle{Batch graph.} Figure~\ref{fig:kernel_bs1024_k64} shows the speedup and bandwidth utilization of \name{} compared to other baselines on the batch graph datasets. For kernel time speedup, \name{} achieves the best performance across all datasets, with an average kernel time speedup of $3.7\times$ (and up to $7.0\times$) compared to DGL sparse across GT, AGNN, and GAT models. Meanwhile, cuGraph and dgNN baselines achieve an average speedup of $2.4\times$ and $1.7\times$, respectively. The \name{} speedup ratio increases with graph degrees, particularly evident in the PATTERN and CLUSTER datasets with high average degrees. This demonstrates that \name{} can effectively handle batch graphs with varying degrees, maintaining optimal kernel performance even as the kernel memory access demands rise. Meanwhile, since AT-GNNs kernels are typically memory-bound, memory bandwidth is a crucial determinant of kernel performance. Therefore, we also analysis the bandwidth utilization of different methods, which is calculated as follows: $\frac{\text{Number of read/write bytes}}{\text{Memory bandwidth} * \text{Kernel execution time}}$. Due to the increase in memory access, the bandwidth utilization ratio decreases for all methods as the graph's average degree increases. Compared to all baselines, \name{} exhibits the highest bandwidth utilization performance across all datasets, demonstrating its advantages from the perspective of memory access.

\stitle{Full graph.}  Figure~\hyperref[fig:full-graph-perf]{10a} shows the speedup on the full graph datasets, including the Citeseer, Cora and Pumbed datasets. Overall, \name{} achieves an average speedup of $16.3\times$ compared to DGL sparse. Meanwhile, cuGraph and dgNN baselines achieve average speedups of $8.3\times$ and $11.7\times$, respectively. \name{} outperforms the other baselines on all datasets, demonstrating notable robustness. Different from batch graph datasets, \name{}, cuGraph, and dgNN achieve higher speedups on full graph datasets. This is because these three graphs are smaller than batch graphs, leading to greater performance gains from kernel fusion. Meanwhile, \name{} and dgNN outperform cuGraph due to the cuGraph framework's more complex call stack, which leads to higher overhead for launching CUDA kernels that cannot be hidden with small workloads.

\stitle{Full graph with super nodes.} Figure~\hyperref[fig:full-graph-perf]{10b} illustrates the speedup on the full graph datasets with super nodes, including the Ogbg-ppa, Reddit and Protein datasets. Because of the presence of super nodes, \name{} employs the \parfusedabb{} method on GT and AGNN convolution, utilizing edge parallel SDDMM to address the performance bottleneck in SDDMM. Among all methods, \name{} consistently outperforms the other baselines and achieves an average $2.8\times$ speedup compared to DGL sparse. 
% Meanwhile, cuGraph and dgNN baselines achieve average speedups of $1.8\times$ and $2.1\times$, respectively. 
Additionally, due to the large number of edges in these graphs, the PyG baseline runs out of memory and is not shown. Compared to Figure~\hyperref[fig:full-graph-perf]{10a}, the speedup of \name{} is reduced because graphs with super nodes are generally larger and exhibit lower kernel launch overhead, diminishing the performance gain from kernel fusion.

\stitle{Ablation Study.}
Ablation experiments in Appendix~\ref{exp:ablation} analyze the performance benefits of each design in \name{} and evaluate its robustness across varying feature dimensions, batch sizes and GPU architectures.

\subsection{End-to-End Performance}\label{exp:endtoend}

\begin{table}[h]
\vspace{-0.3cm}
\centering
\scriptsize
\caption{The E2E time consumption and  speedup of \name{} compared to DGL}
\label{e2e_perf_table}
\begin{tabular}{c|cccc|ccc}
\hline
\multirow{2}{*}{Time (ms)}& \multicolumn{4}{c|}{Batch graph} & \multicolumn{3}{c}{Full graph} \\
& PN  & CL  & PV  & CO & CR & CS & PB \\ \hline
Preprocess &  \textbf{2.01} (1.20)&\textbf{1.81} (1.08)  &\textbf{1.59} (0.86)  &  \textbf{1.61} (0.85)  & -  & - & - \\ 
Forward &  \textbf{2.16} (7.86) &\textbf{1.94} (7.90)  &\textbf{3.23} (8.40)  &  \textbf{3.23} (8.64)  & \textbf{1.99} (7.61)  & \textbf{2.02} (7.75)& \textbf{1.97} (7.69)\\ 
Backward &  \textbf{9.08} (14.9)&\textbf{8.72} (14.2)  &\textbf{8.83} (16.1)  &  \textbf{9.30} (16.4)  & \textbf{3.87} (8.33)  & \textbf{3.89} (8.40)& \textbf{4.89} (8.28)\\
Sum &  \textbf{13.3} (24.0)&\textbf{12.5} (23.2)  &\textbf{13.7} (25.4)  &  \textbf{14.1} (26.0)  & \textbf{5.86} (15.9)  & \textbf{5.91} (16.2)& \textbf{6.86} (16.0)\\ \hline
E2E Speedup &  \textbf{1.81x} & \textbf{1.86x}  & \textbf{1.85x}  &  \textbf{1.83x}  & \textbf{2.72x}  & \textbf{2.73x} & \textbf{2.33x}\\ \hline

\end{tabular}
\end{table}

Table~\ref{e2e_perf_table} presents the E2E speedup of \name{} in comparison to DGL. For batch graph datasets, achieves an average speedup of $1.84\times$ compared to DGL. However, the preprocess step takes slightly longer for \name{} due to the additional conversion to CSR + COO hybrid format for the forward pass and CSC format for the backward pass, whereas DGL only requires the COO format. In the forward and backward pass, \name{} attains average speedups of $3.2\times$ and $1.7\times$ respectively compared to DGL. The lower speedup in the backward pass is attributed to the inability to fuse all backward kernels. For full graph datasets, \name{} outperforms DGL with an average speedup of $2.6\times$ per training epoch, achieving $3.9\times$ speedup in the forward pass and $2.0\times$ in the backward pass. Similar to its kernel performance, \name{} achieves high speedup on full graph datasets because small workloads result in greater performance gains by eliminating kernel launch overhead.
\section{Conclusion}
In this paper, we present \name{}, a framework designed for the efficient execution of AT-GNNs on GPUs. \name{} introduces dynamic bi-level thread scheduling and designs tailored thread scheduling for different operations within AT-GNNs, ensuring the flexibility and generality of fusion strategies. To enhance performance on graphs with super nodes, \name{} designs two general kernel fusion methods, \allfusedabb{} and \parfusedabb{}, which dynamically select appropriate parallel strategies for the SDDMM operation. Additionally, \name{} enables kernel fusion in the backward pass of AT-GNNs, further accelerating end-to-end AT-GNN training. \name{} also integrates with the PyTorch framework and provides APIs to enhance the user experience. Extensive experiments demonstrate that \name{} significantly outperforms existing state-of-the-art GNN frameworks across diverse GNN models and datasets. Furthermore, although \name{} concentrates on single-GPU optimization, its optimization framework can be readily adapted for the training of distributed AT-GNNs, given the consistency of the underlying GPU kernels.

% \section*{Acknowledgements}

\stitle{Acknowledgments} We thank anonymous reviewers for their valuable suggestions. This work is supported in part by the National Natural Science Foundation of China (No. 62222111).

\bibliographystyle{unsrtnat}
\bibliography{reference}
\appendix
\section{Edge Softmax in AT-GNNs}

Here is the vectorized formulation of Softmax operation in AT-GNNs, which normalizes the sparse attention weight matrix $\mathbf{S}$. To enhance the numeric stability of Softmax, the implementation employs a normalization trick, as illustrated in Equation~\eqref{equ:softmax}. It involves two rounds of information exchange between nodes and edges: the first round is to obtain the maximum value $z^{(max)}_v$ from the corresponding row, while the second round sums all connected edge values $f_{uv}$.

\begin{equation}
\begin{aligned}
\text{Round.1: } z^{(max)}_v &= \max_{u\in \mathcal{N}(v)}(s_{uv}) \\
f_{uv} &= \exp(m_{uv} - z^{(max)}_v) \\
\text{Round.2: } z^{(sum)}_v &=\sum_{u\in \mathcal{N}(v)} f_{uv} \\
p_{uv} &= \frac{f_{uv}}{z^{(sum)}_v} \\
\end{aligned}
\label{equ:softmax}
\end{equation}

\section{Backward pass in AT-GNNs}
\label{sec:bwd_at_gnn}
\stitle{Backward pass of SpMM.} The backward pass of SpMM in Equation~\eqref{equ:graph_conv}, as derived from the loss $L$, consists of one SDDMM operation and one SpMM operation, which can be represented by the following equations:
\begin{align}
\frac{\mathrm{d} L}{\mathrm{d} \mathbf{P}}&= \left(\frac{\mathrm{d} L}{\mathrm{d} \mathbf{O}} *\mathbf{V}^T\right)\odot \mathbf{I}\\
\frac{\mathrm{d} L}{\mathrm{d} \mathbf{V}}&= \mathbf{P}^T *\frac{\mathrm{d} L}{\mathrm{d} \mathbf{O}}
\end{align}

\stitle{Backward pass of SDDMM.}
The backward pass of SDDMM in Equation~\eqref{equ:graph_conv}, as derived from the loss $L$, involves two SpMM operations, represented by the following equations:
\begin{align}
\frac{\mathrm{d} L}{\mathrm{d} \mathbf{Q}}&= \frac{\mathrm{d} L}{\mathrm{d} \mathbf{S}}*\mathbf{K} \\
\frac{\mathrm{d} L}{\mathrm{d} \mathbf{K}}&= \frac{\mathrm{d} L}{\mathrm{d} \mathbf{S}}^T*\mathbf{Q}  
\end{align}

\section{Ablation Study}\label{exp:ablation}

\begin{figure}[h]
    \centering
    \includegraphics[width=0.68\linewidth]{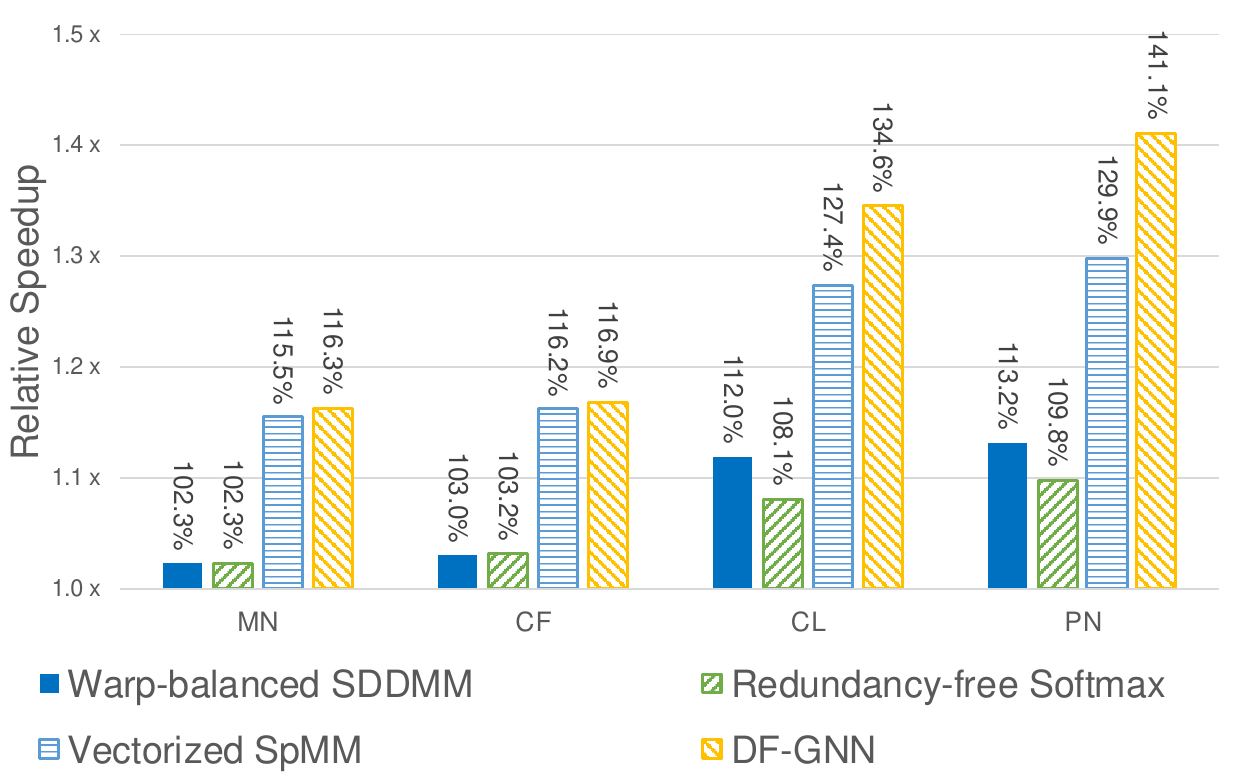}
    \caption{Relative performance improvement brought by proposed optimizations on GT.}
    \label{fig:ablation_intra}
\end{figure}

\stitle{Proposed optimization methods.} Figure \ref{fig:ablation_intra} presents the relative speedup obtained when the three optimization methods proposed in Section~\ref{sec:intra thread schedule}. This demonstrates the impact of the three optimization methods on the final kernel performance. Across all four datasets, the SpMM optimization achieves significant performance improvements, indicating that memory access for SpMM is the primary performance bottleneck in the AT-GNNs kernel. Meanwhile, the optimizations for SDDMM and Softmax provided only a $2\%$ to $3\%$ performance improvement for datasets with low node degrees, such as MNIST and CIFAR10. However, for datasets with high node degrees, such as CLUSTER and PATTERN, these optimizations yield approximately a $10\%$ performance gain. This indicates that the performance gain from the SDDMM and Softmax optimizations depend on the properties of the graph. The greater the number of edges in the graph, the higher the performance gains from these optimizations.

\begin{figure}[h]
    \centering
    \includegraphics[width=0.67\linewidth]{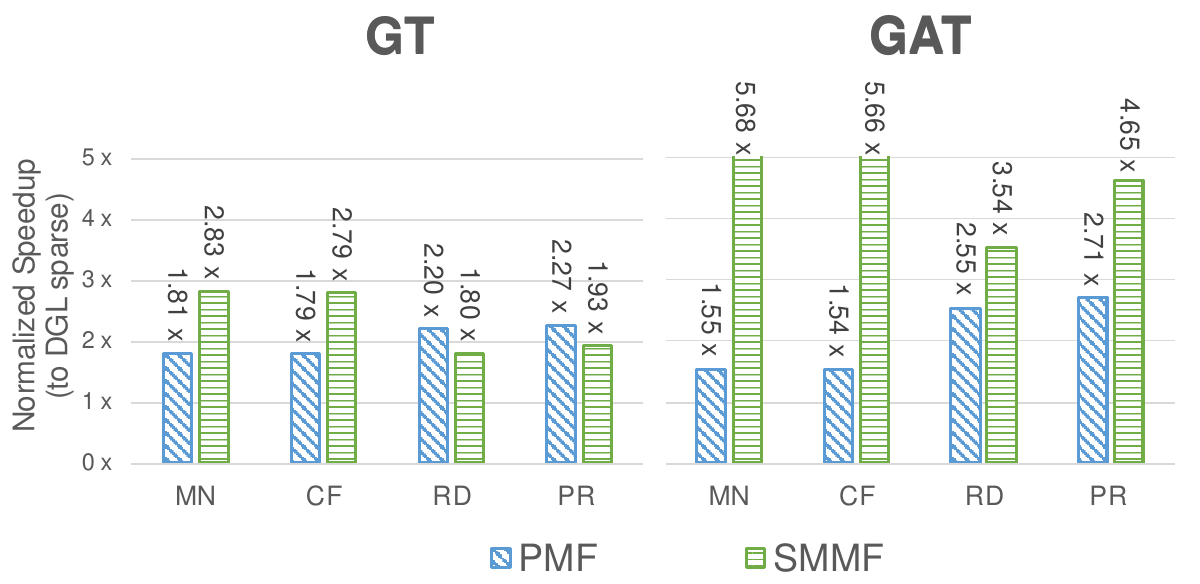}
    \caption{Normalized kernel time speedup of \allfusedabb{} and \parfusedabb{} in \name{}.}
    \label{fig:shmoo_partial_all}
\end{figure}

\stitle{\allfusedabb{} and \parfusedabb{}.} Figure \ref{fig:shmoo_partial_all} shows the speedup of \allfusedabb{} and \parfusedabb{}. For GT convolution with dot-SDDMM, \allfusedabb{} outperforms \parfusedabb{} on the MNIST and CIFAR10 datasets. Meanwhile, on Reddit and Protein datasets with super nodes, \parfusedabb{} shows improved performance compared to \allfusedabb{}. Nonetheless, the incomplete fusion of kernels in \parfusedabb{} leads to only a modest enhancement when compared to \allfusedabb{}. For GAT convolution with add-SDDMM, \allfusedabb{} demonstrates better performance than \parfusedabb{} across all four datasets. This indicates that the \parfusedabb{} method can better handle cases where the input graph contains super nodes and the graph convolution uses dot-SDDMM, demonstrating that the choice between node parallel SDDMM and edge parallel SDDMM in Section \ref{sec:intra thread schedule} is reasonable.

\begin{figure}[h]
    \centering
    \begin{minipage}[t]{.45\textwidth}
        \centering
        \includegraphics[width=\linewidth]{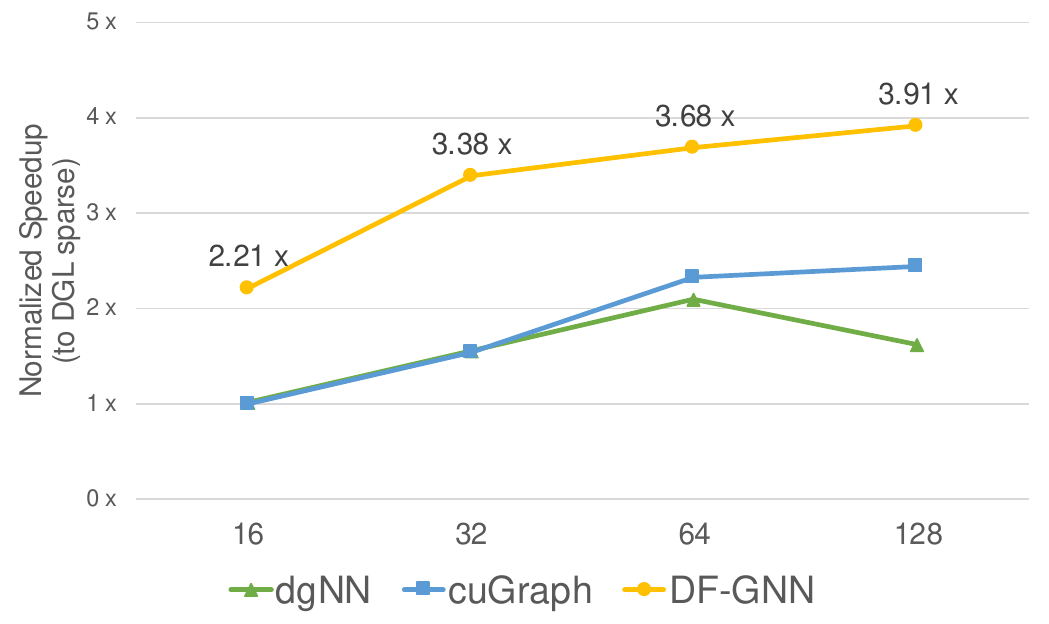}
        \caption{Normalized speedup of GT against different feature dimension (x-axis).}
        \label{fig:shmoo_feature_size}
    \end{minipage}%
    \hspace{0.05\textwidth}
    \begin{minipage}[t]{.45\textwidth}
        \centering
        \includegraphics[width=\linewidth]{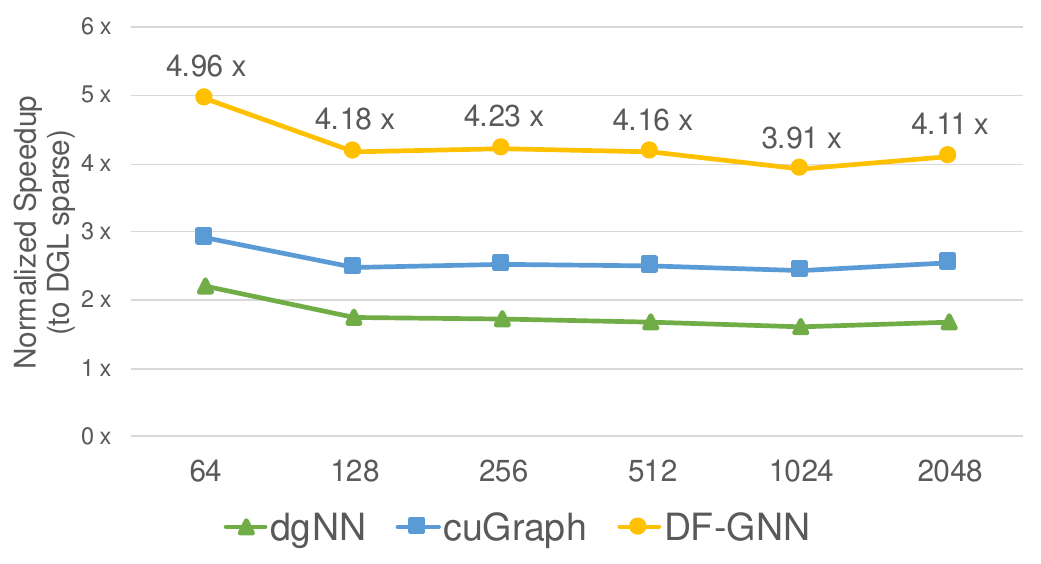}
        \caption{Normalized speedup of GT against different batch sizes (x-axis).}
        \label{fig:shmoo_bs}
    \end{minipage}
\end{figure}

\stitle{Feature Dimension.} Figure \ref{fig:shmoo_feature_size} shows the normalized kernel time speedup as the feature dimension increases from $16$ to $128$ on the PATTERN dataset. \name{} achieves an average of $3.30\times$ speedup, while the dgNN and cuGraph baselines achieve $1.74\times$ and $1.56\times$ speedup, respectively. \name{} consistently achieves better performance compared to the baselines across different feature dimension settings. This demonstrates the extensibility and robustness of \name{} in handling varying feature dimensions. The strong performance of \name{} is further validated by its ability to outperform state-of-the-art methods under diverse feature dimension configurations.

\stitle{Batch Size.} Figure \ref{fig:shmoo_bs} shows the normalized kernel time speedup as the batch size increases from $64$ to $2048$ on the PATTERN dataset. Compared to dgNN and cuGraph, \name{} achieves an average speedup of $4.26\times$ and exhibits superior performance across all batch sizes. This means that the throughput of \name{} can scale proportionally with an increasing batch size, demonstrating its effective handling of graphs with large batch sizes and efficient utilization of GPU resources.

\begin{figure*}[h]
    \centering
    \includegraphics[width=\textwidth]{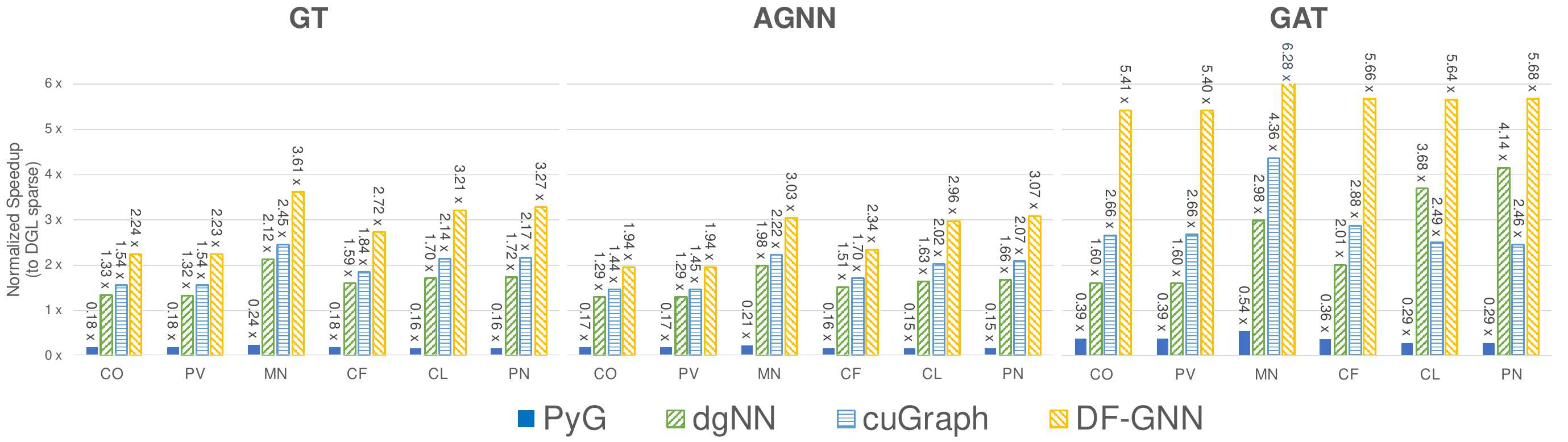}
    \caption{Normalized kernel time speedup on NVIDIA H100 80GB.}
    \label{fig:batch_a100}
\end{figure*}

\stitle{Evaluation on Different GPUs.} The \name{} exhibits robust generalization capabilities, making it applicable across various GPU architectures. Figure \ref{fig:batch_a100} illustrates the normalized kernel time speedup on the NVIDIA H100 80GB GPU. Consistent with the experiment results shown in Figure  \ref{fig:kernel_bs1024_k64}, \name{} outperforms all baselines across all datasets, achieving an average kernel time speedup of $3.81\times$ (and up to $6.28\times$) when compared to DGL sparse across the GT, AGNN and GAT models. This underscores the adaptability of \name{}'s optimization techniques across different GPU architectures, highlighting its potential for widespread applicability in accelerating GNN workloads.

\end{document}